\documentclass[journal]{IEEEtran}

\usepackage{xcolor,soul,framed} 

\usepackage{framed,multirow}
\usepackage{graphicx}

\usepackage{amssymb}
\usepackage{latexsym}

\usepackage{url}
\usepackage{xcolor}

\usepackage{hyperref}

\usepackage{booktabs}

\newcommand{\be}{\begin{equation}}
\newcommand{\ee}{\end{equation}}

\newcommand{\figref}[1]{Fig.~\ref{#1}}
\newcommand{\tabref}[1]{Table~\ref{#1}}

\definecolor{newcolor}{rgb}{.8,.349,.1}

\begin{document}

\title{Stain-invariant self supervised learning for histopathology image analysis}
\author{Alexandre~Tiard$^1$, Alex~Wong$^2$, David~Joon~Ho$^3$, Yangchao~Wu$^1$, Eliram~Nof$^5$, Alvin~C.~Goh$^4$ Stefano~Soatto$^1$, Saad~Nadeem$^4$\\ 

\vspace{5mm}
$^1${Department of Computer Science, UCLA, Los Angeles 90095, USA}\\
$^2${Department of Computer Science, Yale University, New Haven, CT 06511, USA}\\
$^3${Department of Pathology, Memorial Sloan Kettering Cancer Center, New York, NY 10065, USA}\\
$^4${Department of Surgery, Memorial Sloan Kettering Cancer Center, New York, NY 10065, USA}\\
$^5${Department of Medical Physics, Memorial Sloan Kettering Cancer Center, New York, NY 10065, USA}\\

\thanks{Corresponding authors: tiard@cs.ucla.edu and nadeems@mskcc.org.}}

\maketitle

\begin{abstract}
We present a self-supervised algorithm for several classification tasks within hematoxylin and eosin (H\&E) stained images of breast cancer. Our method is robust to stain variations inherent to the histology images acquisition process, which has limited the applicability of automated analysis tools. We address this problem by imposing constraints a learnt latent space which leverages stain normalization techniques during training. At every iteration, we select an image as a normalization target and generate a version of every image in the batch normalized to that target. We minimize the distance between the embeddings that correspond to the same image under different staining variations while maximizing the distance between other samples. We show that our method not only improves robustness to stain variations across multi-center data, but also classification performance through extensive experiments on various normalization targets and methods. Our method achieves the state-of-the-art performance on several publicly available breast cancer datasets ranging from tumor classification (CAMELYON17) and subtyping (BRACS) to HER2 status classification and treatment response prediction.
\end{abstract}


%
\IEEEpeerreviewmaketitle


\section{Introduction}

\IEEEPARstart{T}{he} digitization of tissue samples into Whole Slide Images (WSIs) opens new avenues for computational approaches in histopathology. Hematoxylin and eosin (H\&E) WSIs remain the gold standard for analysis of cancer tissue, but there is an inherent variability in these images due to several factors. Different tissue types or tissues from different patients, scanners and staining protocols will all contribute to introducing undesirable artifacts which create a considerable challenge for the automated analysis of WSIs \cite{Boschman2022-he}. While recent efforts in Computational Pathology (CPath) have 
seen some empirical successes, these methods do not generalize well, which limits their use in a clinical setting \cite{laak2021}. Adapting the method for a specific center requires retraining, which is expensive due to the need for pathologists to manually generate ground truth labels. 

While the underlying methods differ -- from convolutional neural networks \cite{Graham2019-yz,Ho2021-gh}, to graph neural networks \cite{Javed2020-kf,Pati2021-ck}, and multiple instance learning \cite{Lu2020-wn,Quiros2021-yj} formulations -- one constant is the necessity to address the vast stain variations inherent to the slide creation process  \cite{Tellez2019-ur}. 
Earlier works used statistical matching methods \cite{Bautista2014-yi,Kothari2011-ej,Reinhard2001-ts} or a color-deconvolution process, through singular value decomposition (SVD) in \cite{Macenko2009-zp} and Sparse Non-negative Matrix Factorization (SNMF) in \cite{Vahadane2016-gd}. These methods are used to align a set of samples to a specific target, reducing variations in hue, contrast and other staining artifacts, which in turn improves performance of computational models on that set, but at the cost of generalizability to samples that have not been pre-processed by the specific alignment method. Another drawback of these methods is the introduction of spurious artifacts \cite{Wagner2021-ls}. There has been a growing interest in finding ways to make downstream tasks more robust to stain variations.

Many methods have been proposed to address this issue: \cite{Chang2021-tx} use SNMF to perform stain separation, but do so on both a source and target domain to then mix-up the stainings, bridging the domain gap. While this provides robustness to their tumor classification model, it requires access to the target domain, which is not a feasible assumption in practice. Generative adversarial networks (GANs) have been used for style transfer \cite{Tellez2019-ur,Wagner2021-ls} in a pretraining step. \cite{Ke2021-hr} propose a self-supervised clustering pretraining step, with the goal of producing latent input to a normalization GAN. \cite{Nadeem2020-yt} leverage the Wassterstein distance to cast the problem under a Optimal Transport framework.  \cite{Ciga2022-qq} apply SimCLR \cite{chen2020simple} and perform unsupervised pretraining with a stochastic data augmentation strategy on datasets containing multiple organs and staining types. However, as stain normalization is not used directly as augmentation,  the stain variations that their model can handle depends on those present in the data. This is circumvented by pretraining on 57 large scale datasets containing tissues from various organs in order to present the model with enough variability, which is not scalable. 

We instead propose to learn a representation that embeds the same sample under different stainings close to each other in latent space. Accordingly, different images are embedded far apart. This yields a representation that is robust to staining protocols and stain alignment methods. This is unlike previous methods that rely on data pre-processing by shifting the data distribution \cite{Macenko2009-zp,Reinhard2001-ts,Vahadane2016-gd}, or on the assumption that one has access to many large scale datasets whereby the model observes enough variations in staining \cite{Ciga2022-qq,Ke2021-hr,Wagner2021-ls}. In contrast, we avoid adding further steps to the already cumbersome histology pipeline by jointly optimizing the classification objective in our method with a latent-embedding loss. During training, an image is randomly selected from each batch to serve as a normalization target. Other images in the batch are fed through a randomly selected color deconvolution pipeline \cite{Macenko2009-zp,Reinhard2001-ts,Vahadane2016-gd}, yielding a batch of normalized images. We forward both batches of images through a classifier for several downstream tasks - breast cancer classification, subtyping, HER2 status and treatment response prediction - while enforcing that the latent embeddings of images that correspond to the same sample under different stainings are mapped near each other, and far away from other samples. Contrary to contrastive learning methods \cite{Ciga2022-qq,Wagner2021-ls}, this topological constraint is imposed on the original data directly, which creates a stronger structure in the latent space by creating tighter clusters of data under different stain variations. This latent representation of the data 
is robust to stain variations, but it also improves classification performance. By selecting a different target at every batch, we introduce considerable variation without requiring access to several datasets, which reduces the data processing burden on the user.

\textbf{Our contributions:} (i) We propose a latent embedding strategy that yields robust models for breast cancer subtyping, patch level HER2 classification, trastuzumab treatment response and binary tumor classification of raw images from a number of different centers without additional cost of model parameters. (ii) We show that our method alleviates the need for stain normalization customary in existing works. (iii) We show that the topological constraint imposed on our latent space provides a statistically significant improvement in accuracy, achieving state-of-the-art results. (iv) We demonstrate our robustness to stain variations via extensive experiments on the BRACS \cite{Pati2021-ck} dataset.

\section{Method}
We propose a method to infer the cancer subtype class $y \in \mathcal{K} := \{1, 2, ..., K\}$ from an H\&E stained image $x \in \mathbb{R}^{H \times W \times 3}$ where $H$ and $W$ are the height and width of $x$. To achieve this, we leverage a deep neural network $f_{\theta}$, parametrized by $\theta$, that outputs the logits  $f_{\theta}(x) \in \mathbb{R}^{K}$ for an image $x$. The confidence for each class is represented by the softmax response $\sigma(f_{\theta}(x)) \in [0, 1]^{K}$, and to obtain the inferred class, we take the maximum response $\hat{y} = \arg\max(\sigma(f_{\theta}(x)))$.

To train $f_{\theta}$, we assume access to a dataset that is comprised of tuples containing an H\&E stained image $x$ and its corresponding ground truth cancer subtype label $y \in \mathcal{K}$ i.e. $(x, y)$. To obtain a representation that is robust to variations in staining, we propose to learn an embedding that maps an image and its stain-normalized variant close to each other in the latent space through $f^l_{\theta}(x) \in \mathbb{R}^{h \times w \times M}$ where $f^l_\theta$ denotes the $l$-th layer of the network $f_{\theta}$, $h$ and $w$ the height and width of its output, and $M \in \mathbb{N}^*$ the number of channels.

\subsection{Learning Representations Robust to Stain Variations}
To learn a representation that is robust to stain variations, we impose a loss on our latent embedding  to ``pull'' any image and its stain normalized variation close to each other while ``pushing'' them away from other images in an embedding space. This is inspired by contrastive learning, but differs in that we consider the raw image as a cluster center in latent space, and aim to embed its variations close to it. On the other hand, contrastive learning ``pulls'' pairs of \textit{augmented} images close together in the latent space. In our method, for each image within a mini-batch $X \in \mathbb{R}^{B \times H \times W \times 3}$, we select a reference sample $x_{r} \in X$ to be the target of a stain normalization method $g(\cdot)$, for instance \cite{Vahadane2016-gd}, and we normalize the batch $X$ with respect to $x_{r}$ via $\bar{X} = g(X, x_{r}) \in \mathbb{R}^{B \times H \times W \times 3}$ to obtain the stain normalized version of $X$. We now define a positive image pair, or images that should be close in to each other in embedding space, as an H\&E image and its stain normalized variant and the set of $B$ positive pairs $S_p := \{(x_{i}, \bar{x}_{i})  \}$ for $i \in \mathcal{B} := \{1, 2, ..., B\}$, $x_{i} \in X$, and $\bar{x}_{i} \in \bar{X}$. We also define a negative image pair, or images that should be far from other other in embedding space, as any image and any stain normalized variants that do not represent the same sample and their set $S_n := \{(x_{i}, x_{j}) \} \cup \{(x_{i}, \bar{x}_{j})\}$ with cardinality $B^2$ for $j \in \mathcal{B}$ and $j \neq i$ . 

Using the positive and negative sets of image pairs, we aim to minimize the distance between the latent embeddings of images in $S_p$ while maximizing the distance between those in $S_n$. This imposes constraints on the topology of the embedding space during training, unlike previous works for histopathology image analysis \cite{Graham2019-yz,Ho2021-gh,Pati2021-ck}, that allows us to be robust to the color changes  i.e. illumination, saturation, contrast as well as artifacts introduced by the normalization process. To this end, we minimize a self-supervised learning objective $\ell_c$ imposed on the embeddings for both sets of images in each batch. Our self-supervised learning objective or loss function reads:
\begin{equation}
    \ell_c(\theta) = -\sum_{x_{i}, \bar{x}_{i} \in S_p} \log \frac{exp(\texttt{sim}(f^l_{\theta}(x_{i}), f^l_{\theta}(\bar{x}_{i}))/ \tau}{\displaystyle \sum_{x_{i}, x_{j} \in S_n} exp(\texttt{sim}(f^l_{\theta}(x_{i}), f^l_{\theta}(x_{j})) / \tau)},
\end{equation}

where $\texttt{sim}(u, v) = \frac{u . v}{\parallel u \parallel \parallel v \parallel}$ is the cosine similarity between $u$ and $v$, $\tau = 0.70$ the empirically chosen temperature, and with an abuse of notation $x_{j}$ the negative pairing sample. 

To learn the map from an H\&E stained image $x$ to a cancer subtype label $\hat{y}$, we additionally minimize the standard cross entropy loss:
\begin{equation}
    CE(f_\theta(x), \bar{y}) = - \sum_{k=1}^{K} (\bar{y}[k] \log(f_\theta(x)[k])),
\end{equation}
where $\bar{y}$ is the one-hot vector of $y$, and $\bar{y}[k]$ and $f_\theta(x)[k]$ are the $k$-th element in each vector. We impose this loss on both $X$ and its corresponding $\bar{X}$:
\begin{equation}
    \ell_e(\theta) = \frac{1}{2B} \big( \sum_{x_i \in X} CE(f_\theta(x_i), \bar{y}_i) + \sum_{\bar{x}_i \in \bar{X}} CE(f_\theta(\bar{x}_i), \bar{y}_i) \big).
\end{equation}

Our full training objective is the linear combination of our self-supervised learning loss and cross entropy loss, each weighted by scalar hyper-parameter:
\begin{equation}
    \ell(\theta) = w_c \cdot \ell_c(\theta) + w_e \cdot \ell_e(\theta).
\end{equation}
where $w_c = 1.0$ and $w_e = 1.0$ are chosen empirically.

\begin{figure*}[t]
    \includegraphics[width=\linewidth]{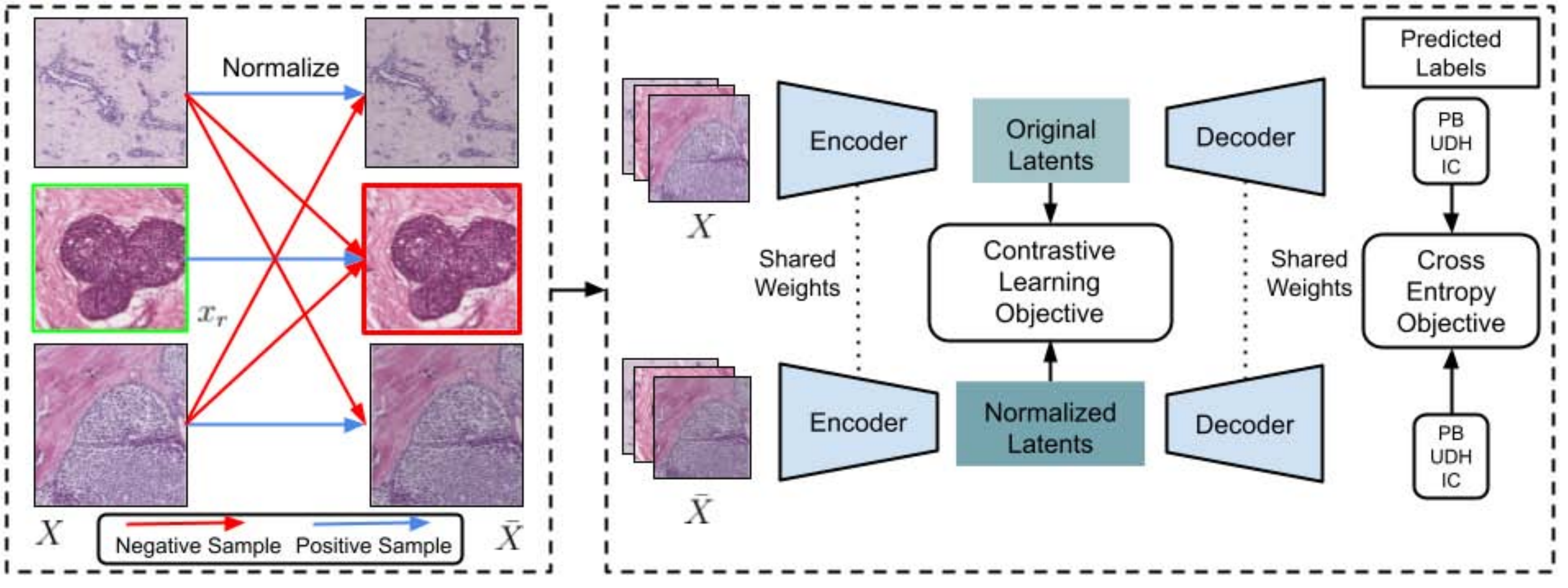}
    \caption{\textit{Training methodology}. (Left) We normalize the raw batch $X$ with respect to a randomly chosen reference image $x_r \in X$ to yield stain-normalized batch $\bar{X}$. We denote a raw image $x$ and its stain-normalized variant $\bar{x}$ as a positive pair and any image pair otherwise as a negative pair. (Right) We feed the raw and stain-normalized images through a deep neural network to obtain their latent embedding. We minimize cross entropy as well as a contrastive objective to learn a representation that maps positive pairs to be close and negative pairs to be far apart in the embedding space.}
\label{fig:pipeline}
\end{figure*}

\section{Experiments}
We demonstrate our method on 3 publicly available datasets: the BReAst Carcinoma Subtyping dataset (BRACS) \cite{Brancati2021-wj}, a HER2 and trastuzumab treatment dataset \cite{farahmand2022deep} and CAMELYON17, a tumor classification dataset \cite{Bandi2019-gy}. We compare against the baselines \cite{farahmand2022deep,he2016deep}, recent \cite{Ciga2022-qq,Wagner2021-ls} and state-of-the-art \cite{Pati2021-ck} methods, as well as a version of our model trained using a contrastive learning framework (which we denote $Ours^{(c)}$). To do so, we transform both batches in Figure \ref{fig:pipeline} through different normalization functions with respect to different targets., instead of conserving a batch of raw data. We outperform all models on several downstream tasks: (i) breast cancer subtyping for (a) images normalized to a specific target sample of \textit{their choosing}, (b) images normalized to different target samples randomly selected over the dataset as well as raw images (without stain normalization); (ii) patch level HER2 classification of raw images from different centers acquired with different scanners; (iii) trastuzumab treatment response in raw images from a single center and (iv) binary tumor classification of raw images from 5 different centers. We also show that using our latent embedding formulation (i) yields tighter clusters in the learnt latent space when compared to contrastive learning and (ii) that the colinearity between stain-normalized versions of the data and its unnormalized version is greater.

\subsection{Implementation details}

\textbf{Network architecture.} We chose a ResNet18 encoder backbone \cite{he2016deep} with 32, 64, 128, 256, 256 filters, in each layer respectively, with leaky ReLU activations. We use two fully connected layers of 256 and 512 neurons to map the latent encoding to logits for the downstream classification task. All weights use random uniform initialization.

\textbf{Training.} We optimize our network using Adam \cite{kingma2014adam} with $\beta_1 = 0.9$ and $\beta_2 = 0.999$. We perform random horizontal and vertical flips, rotations between $[-20^\circ, 20^\circ]$, zero-mean additive gaussian noise with standard deviation of 0.08, and crop sizes up to 90\% of the image height and width as data augmentation. Each augmentation has a 50\% probability of being applied. For every batch, we randomly choose the normalization function $g(\cdot)$ to be applied from \cite{Macenko2009-zp,Reinhard2001-ts,Vahadane2016-gd} with equal probabilities.

\subsection{BRACS} 
BRACS \cite{Brancati2021-wj} is comprised of 547 H\&E WSIs from 189 patients acquired with an Aperio AT2 scanner (40x resolution, $0.25\mu$m/pixel). This dataset was specifically designed to encompass large variability of tissue morphology and staining types. 4537 labeled Regions of Interest (RoIs) are extracted from the WSIs. These RoIs are split between 7 breast subtypes: normal (N), pathological benign (PB), usual ductal hyperplasia (UDH), flat epithelial atypia (FEA), atypical ductal hyperplasia (ADH), dual carcinoma in situ (DCIS), and invasive carcinoma (IC). The authors identified issues with the original splits of the dataset post-publication, and released a second version of the data to address them. We use the second version of BRACS in our work following the original authors' recommendation. The data is split into 3657 RoIs for training, 312 for validation and 570 for testing. The RoIs have variable image dimensions.

\textbf{Test sets.} To evaluate our method on BRACS using the evaluation protocol proposed by HACT-Net \cite{Pati2021-ck}, we preprocessed and normalized the dataset using SNMF with respect to a specific target image $t_H$ that was chosen by \cite{Pati2021-ck}. We denote this version of BRACS as $\mathcal{D}_{H}$ and its test set $\mathcal{T}_{H}$. In order to test for generalization to stain normalization with respect to a different target, we generate another version of test set $\mathcal{T}_{R}$ by normalizing it with respect to a random target in the test set. We also consider the case where no normalization was performed i.e. the Raw dataset. These test sets are shown in first column of \tabref{table:bracs}. Additionally, in \tabref{table:generalization}, we demonstrate the robustness of our method to three different common stain normalization methods \cite{Macenko2009-zp,Reinhard2001-ts,Vahadane2016-gd} across 69 different targets, each taken from a different patient in the test set and show that our method yields small performance deviations across the test set variations.

\textbf{Dataset specific parameters.} We used an initial learning rate of $1 \times 10^{-4}$ and decreased it to $1 \times 10^{-5}$ after 35 epoch for a total of 60 epochs. We use a batch size of 20 and resize each image to $896 \times 896$. Our network is comprised of 13.6M parameters, which is significantly less than competing methods i.e. 127M \cite{Pati2021-ck}, and 28M \cite{Wagner2021-ls}. Training takes approximately 30 hours on two NVIDIA GTX 1080 Ti GPUs.

\begin{figure}[t!]
\begin{center}
\setlength{\tabcolsep}{2pt}
\setlength{\arrayrulewidth}{1pt}
\begin{tabular}{cc}
\includegraphics[width=0.235\textwidth]{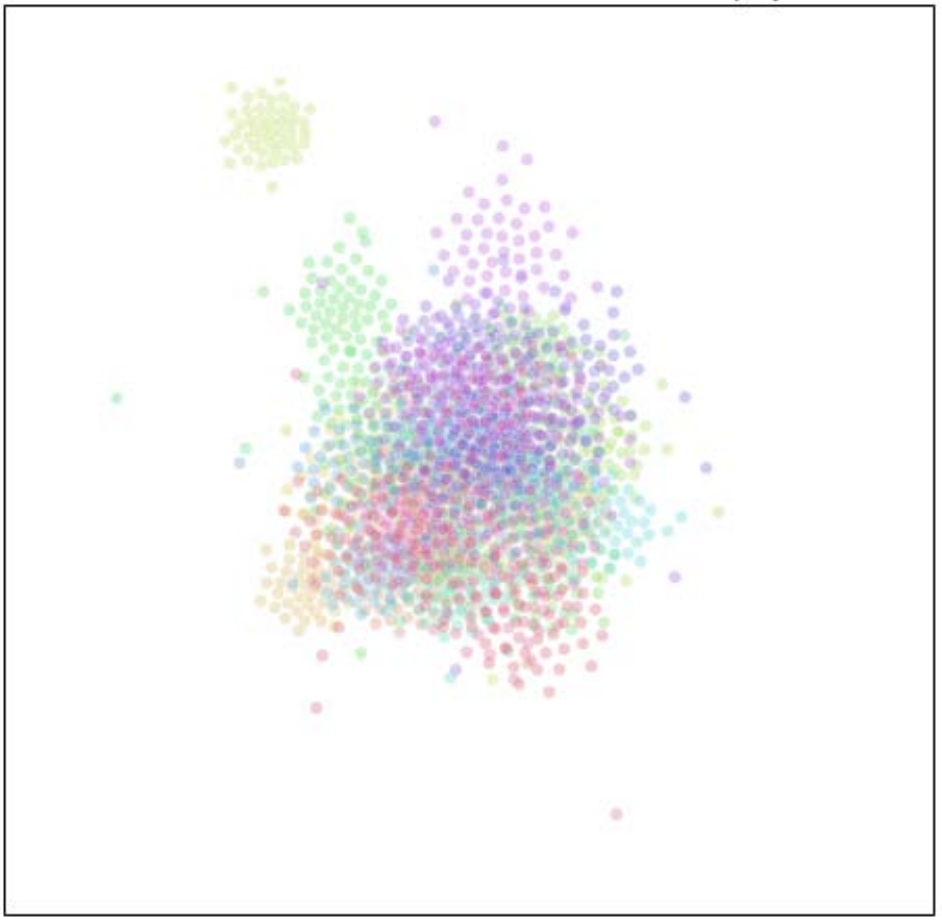}&
\includegraphics[width=0.235\textwidth]{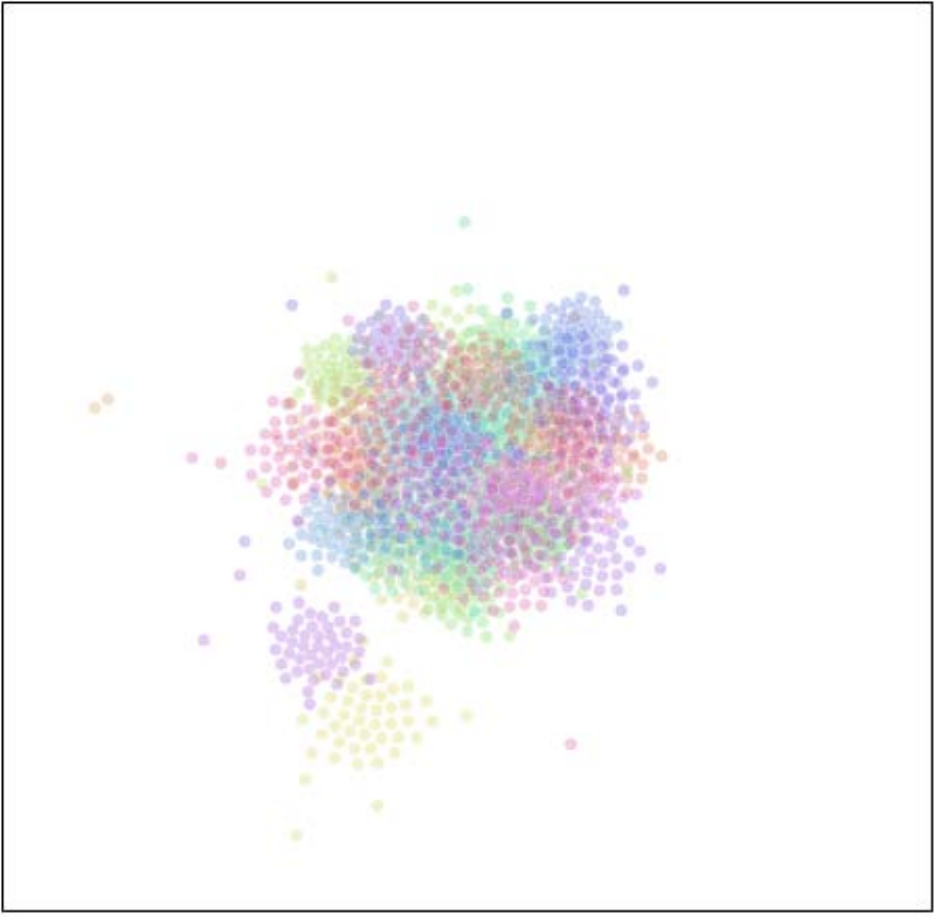}\\
(a) TSNE - Contrastive & (b) TSNE - Ours\\
\end{tabular}
\caption{\textit{Latent space representation}. We use TSNE \cite{van2008visualizing} to plot the latent representation of the latent representation learnt by our network on the BRACS \cite{Brancati2021-wj} dataset using a contrastive learning loss (left) and our latent embedding loss (right). Both plots share the same scale. We assign the same color to latent vectors that correspond to the same sample under different stain variations. Using our latent embedding, we obtain tighter clusters with better separation.}
\label{fig:latent_space}
\end{center}
\end{figure}


\begin{table*}[t]
    \caption{\textit{BRACS test set.} Row 1 lists the datasets used to train each method: Raw denotes the original BRACS dataset, $\mathcal{D}_H$ its normalized version using SNMF \cite{Vahadane2016-gd} and the target chosen by \cite{Pati2021-ck}. Row 2 denotes the mean inference time and row 3 the model size in number of parameters. Rows 4-6 show the mean accuracy on different test sets: $\mathcal{T}_H$ shows comparisons to the state of the art \cite{Pati2021-ck} on data normalized to the target chosen by \cite{Pati2021-ck}, where we outperform \cite{Pati2021-ck} as well as \cite{Ciga2022-qq} who trained on 57 additional datasets. We also outperform a version of our model trained with a contrastive learning approach, where both batch of images are transformed by a random normalization method. $\mathcal{T}_R$ and Raw show comparisons on data normalized to a randomly target sample and data without the customary normalization, respectively, which demonstrate the sensitivity of existing methods to the choice of normalization target or the lack thereof. Our method is robust to stain-normalized images for different targets as well as raw images; whereas other methods experience a drop in performance.} 
    \footnotesize
    \centering
    \setlength\tabcolsep{6pt}
    \begin{tabular}{| l | l | c | c | c | c | c | c | c | c | c |}
        \toprule
        & & HACT & HACT  & Baseline & Baseline & Cigav1 & Cigav2 & HistAuGAN & $Ours^{(c)}$ & Ours \\
        & Train & Raw & $\mathcal{D}_H$ & Raw & $\mathcal{D}_H$ & Raw, & Raw, & Raw, & Raw & Raw \\
        & Set & & & & & 57 others \cite{Ciga2022-qq} & 57 others \cite{Ciga2022-qq} & CAMELYON17 \cite{Bandi2019-gy} & \\
        \cmidrule{2-11}
        & Time & 26s & 26s & 20ms & 20ms & 20ms & 20ms & 20ms & 20ms \\
        \cmidrule{2-11}
        & Size & 127M & 127M & 13M & 13M & 13M & 13M & 28M & 13M & 13M \\
        \midrule
        \midrule
        \multirow{4}{0.6cm}{\parbox{0.6cm}{Test Set}}
        & $\mathcal{T}_H$ & 49.3 & 55.3 & 47.0 & 51.6 & 49.1 & 55.3 & 49.5 & 58.2 & \textbf{60.2} \\
        \cmidrule{2-11}
        & $\mathcal{T}_R$ & 34.0 & 26.0 & 43.2 & 48.8 & 53.3 & 53.4 & 48.4 & 58.4 & \textbf{61.0} \\
        \cmidrule{2-11}
        & Raw & 52.8 & 47.1 & 48.9 & 46.0 & 58.9 & 60.2 & 50.0 & 59.2 & \textbf{61.7} \\
        \midrule
    \end{tabular}
\label{table:bracs}
\end{table*}

\begin{figure*}[t]
    \includegraphics[width=\linewidth]{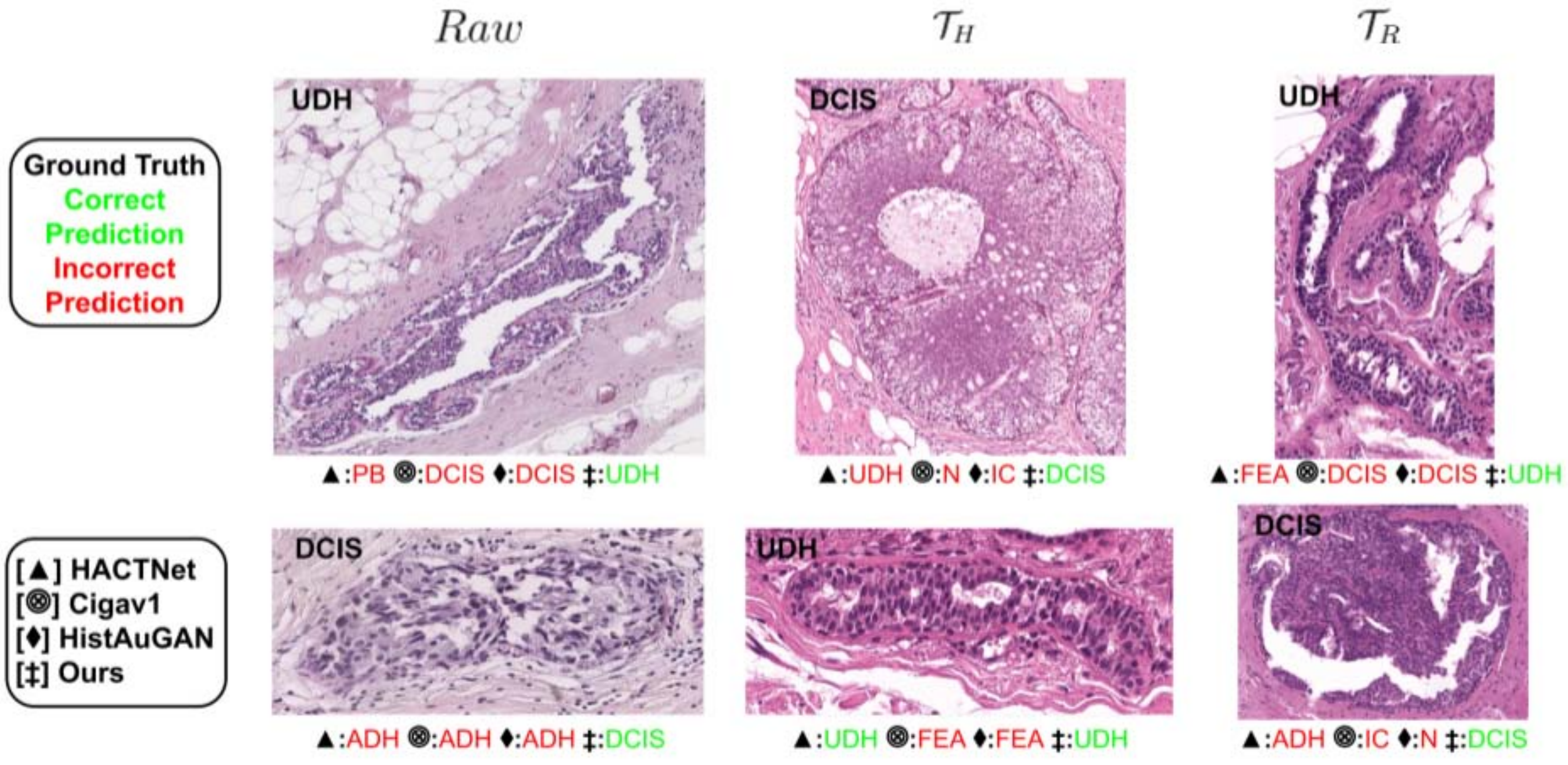}
    \caption{\textit{Sample predictions.} Left column: Raw dataset, our method alone is able to distinguish DCIS from atypical samples without preprocessing the data. Disambiguating ADH from DCIS is a challenging tasks even for pathologists \cite{Brancati2021-wj}. Middle column: $\mathcal{T}_H$  Stain variations impact predictions, a sample is predicted as both N and IC even though these classes differ the most. Our model can correctly classify different types of atypia. Right column: $\mathcal{T}_R$ the stain variability does not impact our model, which is still able to correctly assess DCIS from atypia.}
\label{fig:results}
\end{figure*}

\textbf{Results on a specific normalization target.}
We evaluate our method on the BRACS test set that has been pre-processed and normalized to a specific target image $t_H$ that \cite{Pati2021-ck} chose and reported state of the art results. Here, we consider two versions of HACT-Net \cite{Pati2021-ck} and the baseline \cite{he2016deep}: First, both \cite{he2016deep,Pati2021-ck} are trained on $\mathcal{D}_{H}$ as specified by \cite{Pati2021-ck}; a second model for each is trained on the raw images \textit{without} stain normalization to test their generalization. Ciga \cite{Ciga2022-qq} and HistAuGAN \cite{Wagner2021-ls} were both designed to be invariant to stain normalization so we train them on the raw data without normalization as per their training protocol. For Ciga, we replaced the last fully connected layer to match our number of output classes. For fair comparison, we also train two version of Ciga where Cigav1 denotes finetuning only the last fully connected layer while freezing the rest of the network to preserve the information stored in their weights that was obtained from pretraining on 57 datasets, and Cigav2 denotes finetuning the entire network.

\tabref{table:bracs} shows a comparison between all model on three different test sets: $\mathcal{T}_H$, $\mathcal{T}_R$, and Raw. Row 4 shows results for $\mathcal{T}_H$, where the test set is normalized according to the target chosen by the state of the art HACT-Net \cite{Pati2021-ck}. Our method trained on raw images outperforms HACT-Net by 4.9\% in mean accuracy (8.9\% relative gain) when they are trained on $\mathcal{D}_{H}$. When HACT-Net is trained on raw images, our improvements grows to 10.9\% mean accuracy (22.1\% relative gain). A similar trend is also observed in the baseline when tested on data normalized to a target that it is not trained on. This shows that choosing a specific target for stain normalization during training in fact limits the applicability of the method. This is particularly limiting when one wishes to use a pretrained model out-of-the-box without intricate knowledge of the specific target sample and normalization method that was used to train the model.

We now focus on methods \cite{Ciga2022-qq,Wagner2021-ls} designed to address the covariate shift caused by different targets and stain normalization methods. Ciga \cite{Ciga2022-qq} aims to close the gap by training on diverse (57 datasets) staining protocols while HistAuGAN \cite{Wagner2021-ls} uses a generative model to introduce diverse stain normalization during training. While both generalize better than HACT-Net \cite{Pati2021-ck} and the baseline, they perform worse than the proposed method. Despite not having access to additional datasets, we outperform both version of Ciga \cite{Ciga2022-qq} (Cigav1, Cigav2) by 11.1\% and 8.9\% mean accuracy (22.6\% and 8.9\% relative gain), respectively. Furthermore, we observe a steep drop in performance when Ciga is tested on stain-normalized images, which shows that training on diverse raw H\&E stained images from 57 datasets does not address the large variations and artifacts introduced by stain normalization. We also observe a similar trend in HistAuGAN \cite{Wagner2021-ls} where they consistently produce scores around 50\% mean accuracy range; whereas, our method consistently yields scores around 61\% range.

We empirically validate that our method can learn a representation that is robust to different stain normalization targets in \tabref{table:bracs} by embedding images and its stain-normalized variants close to each other and different images far apart. Moreover, our method can also be applied to raw images without any stain normalization step customary in CPath. This makes our method widely applicable as it enables pathologists to apply our method out-of-the-box without the need to know the specific stain normalization method or the target sample. \figref{fig:results} shows that our method is able to correctly infer breast cancer subtypes from images that existing methods missed. This includes diambiguating ADH subtype from DCIS, which is a challenging tasks even for pathologists \cite{Brancati2021-wj}.

\begin{table*}[t]
\caption{\textit{Mean and standard deviation of accuracies over different versions of the BRACS test set.} We generate 69 versions of the test set, each normalized by a different method and to a different target. We also include the raw dataset to yield 70 test sets (BRACS70) to evaluate robustness to various stain normalization methods and targets. Our method boasts the highest mean accuracy and lowest deviation across BRACS70.}
    \centering
    \setlength\tabcolsep{10pt}
    \begin{tabular}{l l c c c c c c}
    \toprule
    & & Baseline & Cigav1 & Cigav2 & HGAN & $Ours^{(c)}$ & Ours \\
    \midrule
    & BRACS70 & 46.7 $\pm$ 7.5 & 45.7 $\pm$ 9.6 &  46.6 $\pm$ 8.9 & 45.1 $\pm$ 4.2 & 53 $\pm$ 3.8 & 59 $\pm$ 1.6 \\
    \midrule
    \end{tabular}
\label{table:generalization}
\end{table*}

\textbf{Results on diverse normalization targets and methods.} 
To evaluate the robustness of our method -- not only on stain normalization targets, but also stain normalization methods -- we generate different versions of the BRACS testing with 69 random targets, each selected from a different patient. Each target is randomly assigned to one of three groups of equal size to be normalized by one of three different stain normalization methods: Vahadane \cite{Vahadane2016-gd}, Macenko \cite{Macenko2009-zp}, and Reinhard \cite{Reinhard2001-ts}. We additionally include the raw test images without normalization to yield 70 test sets (BRACS70). For this experiment, we consider the baseline and methods \cite{Ciga2022-qq,Wagner2021-ls} that have generalized best in \tabref{table:bracs}.

\tabref{table:generalization} shows that our method is the most robust. The baseline (without our contrastive objective) achieves 46.7 $\pm$ 7.5, i.e. large variations in performance for different targets and normalization method. Similarly, for Cigav1 and Cigav1 \cite{Ciga2022-qq}, we observe high deviations in accuracies: 46.7 $\pm$ 7.5 and 45.7 $\pm$ 9.6, respectively. This again demonstrates that the stain normalization variability cannot be model simply by training on more data, but require constraints on the representation. We note that amongst the competing methods, HistAuGAN \cite{Wagner2021-ls} achieves the lowest deviation, 45.1 $\pm$ 4.2, but suffers a 5\% mean accuracy drop as compared to \tabref{table:bracs}. In contrast, the proposed method boasts a 59.0 $\pm$ 1.6, which is significantly higher in accuracy with lesser deviation than competing methods. 

\textbf{Comparison to contrastive learning.} To understand why a model trained under our proposed latent embedding outperforms the same model trained under a contrastive learning framework, we look at the learnt latent space. Figure \ref{fig:latent_space} contains a 2-dimensional plot of the embedded vectors within BRACS70 obtained through TSNE \cite{van2008visualizing}, where the embeddings of stain variations of the same image are assigned the same color. Three differences can contribute to the observed improvement in performance. First, clusters are tighter under our learning framework (a $15\%$ drop in mean intra-cluster distance over BRACS70), so stain variations cause a smaller perturbation in the latent space on average, which leads to more stable classification results. Second, cluster centroids are closer to the embedding of the raw data in our framework (a $7\%$ difference with contrastive learning averaged over BRACS70), since the raw data is actually forwarded through our model. This pushes the raw data away from decision boundaries. Third, the colinearity between the raw data and its stain normalized variants is higher under our latent embedding (a $12\%$ difference across BRACS70), which also contributes to the increased robustness.

\subsection{HER2 status and treatment response} 
We denote by HER2 \cite{farahmand2022deep} the agglomeration of 3 cohorts. First, the \textit{Yale HER2 Cohort} (Y-HER2) contains 192 H\&E WSIs scanned using Vectra Polaris by Perkin-Elmer at 20× magnification at Brady Memorial Laboratory Rimm’s lab. The cohort is split in 93 HER2 positive and 99 HER2 negative samples. Second, the \textit{TCGA HER2 cohort} (denoted T-HER2) contains 182 H\&E WSIs (92 HER2 positive and 90 HER2 negative) downloaded from TCGA, the quality of which were assessed by pathologists. Third, the \textit{Yale trastuzumab response cohort} (denoted TRAS) contains 85 samples downloaded from the Yale Pathology electronic database, of which 36 are responders and 49 non-responders. ROIs of variable size, annotated through Aperio ImageScope software, were provided for all cohorts.

\begin{table*}[t]
    \caption{\textit{HER2 status and treatment response.} Row 1 lists the datasets used to train each method, where the supersets indicates which tasks the datasets were used for, if not both. Row 2 denotes the mean inference time and row 3 the model size in number of parameters. Row 4 show the mean accuracy on the TCGA cohort digitized in a different center, where we outperform all methods. Row 5 shows the mean accuracy on the trastuzumab treatment outcome prediciction task.} 
    \scriptsize
    \centering
    \setlength\tabcolsep{1pt}
    \begin{tabular}{| l | l | c | c | c | c | c | c |}
        \toprule
        & & Inceptionv3 & Baseline & Cigav1 & Cigav2 & HistAuGAN & Ours \\
        & Train & $Y-HER2^{1}/TRAS^{2}$ & $Y-HER2^{1}/TRAS^{2}$ & $Y-HER2^{1}/TRAS^{2}$, & $Y-HER2^{1}/TRAS^{2}$, & $Y-HER2^{1}/TRAS^{2}$ & $Y-HER2^{1}/TRAS^{2}$ \\
        & Set & & & 57 others & 57 others & CAMELYON17 & \\
        \cmidrule{2-8}
        & Time & 23s & 20ms & 20ms & 20ms & 20ms & 20ms \\
        \cmidrule{2-8}
        & Size & 23M & 13M & 13M & 13M & 28M & 13M \\
        \midrule
        \midrule
        \multirow{2}{0.6cm}{\parbox{0.6cm}{Test Set}}
        & $T-HER2^{1}$ & 65.0 & 59.7 & 51.8 & 60.3 & 61.2 & \textbf{67.7} \\
        \cmidrule{2-8}
        & $TRAS^{2}$  & 73.0 & 66.0 & 64.3 & 74.3 & 73.3 & \textbf{74.6} \\
        \midrule
    \end{tabular}
\label{table:her2}
\end{table*}

HER2 positive breast cancers tend to be more aggressive than other types of breast cancer, and have a higher likelihood of recurrence \cite{farahmand2022deep}. Specific treatments though anti-HER2 agents are available to improve patient prognosis, but detecting the HER2 status of a metastatic cancer traditionally requires other, more specific imaging modalities such as immunohistochemistry (IHC). Since H\&E imaging is the gold standard for cancer analysis, developing an automated method to infer HER2 status without requiring additional measurements would simplify clinical processes. Table \ref{table:her2} presents results on 2 tasks linked to HER2 status from H\&E images. The first task is HER2 status prediction, where the training and testing set are extracted from different centers and acquired with different scanners. The second task is to predict the response of HER2 positive samples to a specific treatment, trastuzumab. We compare to an Inceptionv3 baseline \cite{farahmand2022deep}, along with other previous methods that aim to address stain variations \cite{Ciga2022-qq,Wagner2021-ls}.

The first task allows us to demonstrate that the representation learned through our model is robust to stains accross centers and acquisition device, as we outperform all others. This is done without requiring access to data from different domains as in \cite{Wagner2021-ls}. Our method also reaches state-of-the-art on the trastuzumab response prediction, which provides actionable information on successful treatment plans.

\begin{table*}[t]
    \caption{\textit{CAMELYON17 binary tumor classification.} Row 1 lists the datasets used to train each method, which contain images from a single center. Row 2 denotes the mean inference time and row 3 the model size in number of parameters. Row 4 show the mean accuracy on test sets, comprised of images from the 4 centers not seen during training.} 
    \scriptsize
    \centering
    \setlength\tabcolsep{1.2pt}
    \begin{tabular}{l c c c c c | c c c c c}
        \toprule
        & & & HistAuGAN & & & & & Ours  & & \\
        Train & $\mathcal{D}_{RUMC}$ & $\mathcal{D}_{CWZ}$ & $\mathcal{D}_{UMCU}$ & $\mathcal{D}_{RST}$ & $\mathcal{D}_{LPON}$ & $\mathcal{D}_{RUMC}$ & $\mathcal{D}_{CWZ}$ & $\mathcal{D}_{UMCU}$ & $\mathcal{D}_{RST}$ & $\mathcal{D}_{LPON}$ \\
        Set & & & & & & & &  \\
        \midrule
        Time & & & 12ms & & & & & 12ms  & & \\
        \midrule
        Size & & & 28M & & & & & 13M & & \\
        \midrule
        \midrule
        \multirow{2}{0.6cm}{\parbox{0.6cm}{Test Set}}
        & $\mathcal{T}_{RUMC}$ & $\mathcal{T}_{CWZ}$ & $\mathcal{T}_{UMCU}$ & $\mathcal{T}_{RST}$ & $\mathcal{T}_{LPON}$ & $\mathcal{T}_{RUMC}$ & $\mathcal{T}_{CWZ}$ & $\mathcal{T}_{UMCU}$ & $\mathcal{T}_{RST}$ & $\mathcal{T}_{LPON}$ \\
        & 92.6 & 57.6 & 61.0 & 95.9 & 84.0 & 92.6 & 64.4 & 63.9 & 96.1 & 88.3
        \\
        \midrule
    \end{tabular}
\label{table:camelyon17}
\end{table*}

\subsection{CAMELYON17} 
CAMELYON17 \cite{Bandi2019-gy} is a multi-center dataset of H\&E stained WSIs that were acquired using different scanners. We denote the 5 medical centers as RUMC, CWZ, UMCU, RST and LPON. 10 WSIs are provided from each medical center, containing pixel-level binary annotations for tumor classification. We tile the full resolution WSIs into $512 \times 512$ pixel images, and assign a tile-level label of 1 for tiles containing over $1\%$ of tumor pixels, 0 otherwise. While the tumor tiles represent a small proportion of the dataset ($7\%$ of all tiles), this proportion is constant across centers. To test the robustness of our representation to variations in staining, we train our model independently on data from each center and perform inference on unseen data from other centers. We refer to $\mathcal{D}_{c}$ as the training set containing images and labels from the center $t \in \mathcal{C} := \lbrace RUMC, CWZ, UMCU, RST, LPON \rbrace$ and $\mathcal{T}_{c}$ as the testing set containing images and labels from the centers $t \in \mathcal{C} \setminus {c}$.

Table \ref{table:camelyon17} contains the results of the binary classification task for HistAuGAN \cite{Wagner2021-ls} and our model. The work of Ciga \cite{Ciga2022-qq} was excluded as their model was trained on 57 datasets which include the five centers from CAMELYON17. We outperfrom HistAuGAN on all centers, while using a single training pipeline and 15M fewer parameters.

\section{Discussion}
We have proposed a method to learn a representation that is robust to stain variations in raw and normalized H\&E images of breast cancer for several classification tasks. Our method imposes constraints on the topology of the latent space through an explicit loss function such that variations of the same image map to a tight cluster. This enables our network, a simple ResNet18, to be more robust than methods that train on 57 datasets \cite{Ciga2022-qq} and that use complex architectures like GANs \cite{Wagner2021-ls}. This is demonstrated by generalizing to a test set acquired in from different centers for the task of HER2 status prediction and tumor classification. This latent space is conducive for several downstream tasks, such as cancer subtyping as we beat the state of the art \cite{Pati2021-ck} on data that is normalized to their choice of target. Our latent embedding loss function allows for faster training than a contrastive learning framework and yields tighter clusters in the latent space and better classification results. We tackle a wide range of tasks in histopathology image analysis, which forms a pipeline from cancer detection and subtyping, to treatment planning based on HER2 status.

However, there are drawbacks as our method can be costly in training: The memory footprint of the loss scales quadratically with respect to batch size when forming the set of negative pairs. This can be mitigated by randomly sampling a fixed number of negative pairs, but doing so may come at the cost of robustness. Our method does have failure modes as discussed in the Supp. Mat. Furthermore, like all blackbox models, our method is not interpretable, so future work may consider incorporating explainability measures to shed light on the morphological indicators of HER2 status within H\&E images and differences between subtypes. Additionally, while softmax scores can serve as a confidence, it is not a true calibrated measure of uncertainty, which is critical for clinical diagnosis -- we leave this direction for future work as well. 

\section*{Acknowledgments}
This project was supported by MSK Cancer Center Support Grant/Core Grant (P30 CA008748).


%





\ifCLASSOPTIONcaptionsoff
  \newpage
\fi






\section*{Supplementary Material}
\noindent \textbf{Code Versioning}
Our method was trained on a machine running Ubuntu 20.04, using Python 3.7, Pytorch 1.7.1 and CUDA 11.4. Library versions will be made available upon code release. Code for HACTNet can be found at \url{https://github.com/histocartography/hact-net}. HistAuGAN weights were obtained from \url{https://github.com/sophiajw/HistAuGAN}. We then trained our baseline model with all augmentations, adding HistAuGAN with a probability 1. The weights for the model denoted Ciga were obtained from \url{https://github.com/ozanciga/self-supervised-histopathology}. \\

\noindent \textbf{Dataset generation}
The BRACS dataset is available at \url{https://www.bracs.icar.cnr.it}. The target used to generate $\mathcal{D}_H$ and $\mathcal{T}_H$ is located at \url{https://github.com/histocartography/hact-net}. See Table \ref{table:test_targets} for selected targets and their corresponding normalization method. For $\mathcal{T}_R$, the randomly selected target is 1823\_DCIS\_3.

\begin{table}[h]
	\scriptsize
	\centering
	\setlength\tabcolsep{1pt}
	\caption{\textit{Filenames of targets selected for BRACS70.} We randomly select a sample for each of the 69 patients in the test set (IDs are included in the filenames) and run via Vahadane, Macenko, and Reinhard stain normalization methods.}
	\begin{tabular}{l l l}
		\toprule
		\textbf{Vahadane} & \textbf{Macenko} & \textbf{Reinhard}\\
		\midrule
		1330\_PB\_1, 1850\_N\_3 &  1824\_UDH\_1, 1940\_PB\_1&  1955\_N\_4, 1283\_FEA\_2\\
		1941\_PB\_4, 1878\_PB\_1 &  1952\_PB\_1, 264\_N\_1 & 292\_N\_2, 1856\_FEA\_2\\
		1814\_PB\_1, 1826\_ADH\_2 &  286\_N\_9, 1906\_PB\_1 &  1943\_PB\_2, 1945\_N\_3 \\
		1599\_N\_1, 1580\_ADH\_1&  1942\_N\_1, 1936\_N\_4 &  1476\_UDH\_2 , 1598\_N\_1\\
		1851\_N\_1, 1849\_UDH\_2  & 1589\_UDH\_1, 1910\_ADH\_3 &  1954\_N\_2, 1848\_UDH\_1 \\
		1844\_ADH\_1, 301\_N\_3 &  1908\_UDH\_3, 1938\_UDH\_4 & 1817\_ADH\_3, 1821\_UDH\_1\\
		1503\_PB\_4, 1601\_N\_1 &  3323\_IC\_1, 1892\_UDH\_2 &  1896\_N\_2, 1825\_UDH\_2\\
		1895\_PB\_3, 305\_N\_1  &  1852\_N\_1, 1416\_UDH\_1&  1872\_PB\_2, 1284\_UDH\_4\\
		1855\_N\_1, 1619\_N\_4 & 1853\_N\_1, 1870\_PB\_1 & 1228\_UDH\_2, 1286\_N\_40 \\
		1619\_N\_4, 310\_N\_3 & 1602\_UDH\_4, 1334\_PB\_1  & 1597\_DCIS\_1, 1590\_DCIS\_1\\
		300\_N\_2, 1897\_PB\_3 & 1937\_N\_6, 1944\_PB\_4  &  1413\_IC\_1, 1823\_ADH\_2\\
		1003694\_ADH\_1 & 1813\_FEA\_1 & 1003691\_ADH\_1\\
		\midrule
	\end{tabular}
	\label{table:test_targets}
\end{table}

\begin{figure*}[t!]
    \centering
    \includegraphics[width=0.9\linewidth]{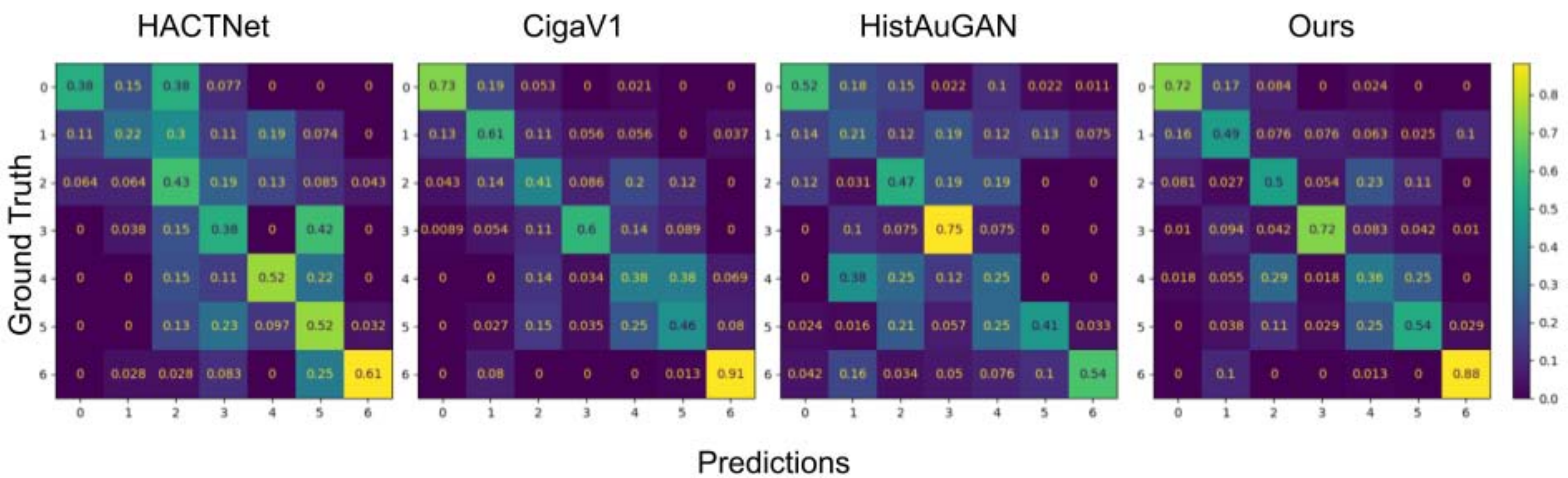}
    \caption{\textit{Confusion Matrices on Raw}}
\label{fig:confusion}
\end{figure*}

\begin{figure*}[t!]
    \centering
    \includegraphics[width=0.9\linewidth]{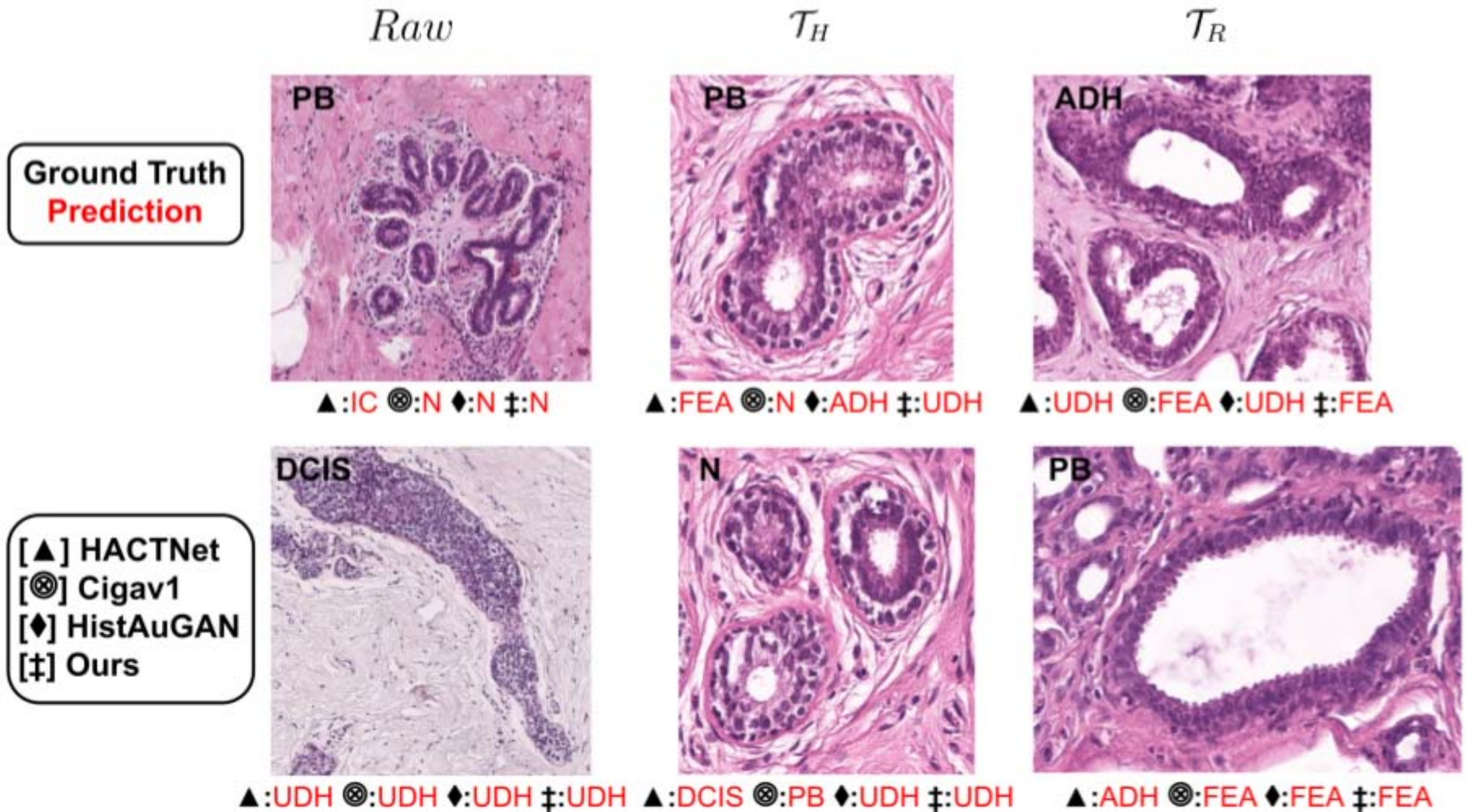}
    \caption{\textit{Failure Cases} Bottom left: non-cancerous prediction of DCIS.}
\label{fig:failure}
\end{figure*}

\noindent \textbf{Failure Modes} Figure \ref{fig:failure} shows failure modes of the tested models. We note that all models fail on these samples. We tend to predict classes within the same superclass i.e. non-cancerous, pre-cancerous or cancerous -- suggesting that granularity of the class labels plays a role. Figure \ref{fig:confusion} presents the confusion matrices on the Raw test set. Disambiguating ADH and DCIS is inherently challenging -- there is high inter-observer variability amongst pathologists when labeling, but our method gets the highest recall for DCIS and on the cancer superclass. \\

\noindent \textbf{Additional Results and Cross-Stain Consistency}
A model prediction is cross-stain consistent if its label for a given sample normalized to different targets stays the same. Across Raw, $\mathcal{T}_H$ and $\mathcal{T}_R$, we are cross-stain consistent for $88$\% of samples, compared to $64$\% for Cigav1, $48$\% for HistAuGAN, and $10$\% for HACTNet. This shows that changing the stain normalization target has a limited impact on our model. This also holds across normalization methods: 80\% of test set data was cross-stain consistent across at least 65 versions of the test sets in $BRACS70$, and 70\% of those samples were correct predictions. \\

\vfill


\end{document}